# Symposium: Trust via Auditable Records for Communities of AI Scientist Agents

Dexter Pratt

Department of Medicine, University of California San Diego
La Jolla, CA 92093, United States
Email: depratt@health.ucsd.edu

## Abstract

Symposium is a formal framework and practical implementation to record the operation of AI agents deployed by small scientific research communities. Symposium provides long-term, immutable histories of agent-driven research activity, leaving auditable trails of analyses, hypotheses, data, and scientific discourse. This shared record of published artifacts enables agents to build on prior work and preserves the evidence researchers and agents need to make purpose-dependent trust assessments. Symposium captures scientific argument, including structured claims, fine-grained evidence citations, assumptions, and explicit declarations of what material may and may not be used as evidence. Symposium differs from AI co-scientist agents or integrated AI research environments; it is a framework that separates a scientific community's durable history from the agents and other systems that operate on that history. It assumes that a community will use diverse AI systems in a rapidly evolving environment. A working implementation of the publication infrastructure, agent prompt components, and documentation are provided to enable users to rapidly set up and run their own Symposium community.

## Introduction

This paper introduces Symposium, a formal framework and practical implementation to record the operation of AI agents deployed by small scientific research communities. Symposium provides long-term, immutable histories of agent-driven research activity, leaving auditable trails of analyses, hypotheses, data, and scientific discourse. This shared record enables agents to build on prior work and researchers to assess trust in the products.

Symposium differs from AI co-scientist agents[1–4] or integrated AI research environments[5]; it is a framework that separates a scientific community's durable history from the agents and other systems that operate on that shared history. It assumes that a community will use diverse AI systems in a rapidly evolving environment.

The first motivation of Symposium is the need to assess trust in agent output. AI agents can increasingly perform aspects of scientific research, such as literature analysis, experiment planning, data analysis, and hypothesis generation. They can perform some tasks far faster than human researchers, potentially accelerating scientific progress. But at the same time, agents are imperfect. They can fabricate facts and citations and misread genuine sources. Just as concerning, they can produce hypotheses that are persuasive but obscure poor reasoning, assumptions, or selection of supporting data. To reap the benefits of rapid AI science, we therefore need mechanisms to assess how much we should trust a given AI-generated output. The answer to this challenge is not simply "build better agents"; better agents can still fail, possibly when the stakes are high.

Agent misalignment is another challenge to trust in AI science, an emerging issue as general progress also makes agents increasingly capable of deception and conspiracy. Advanced models can pursue goals their operators did not set and misrepresent their own behavior[6,7]; prominent examples in cybersecurity are current events as of this writing[8,9]. Researchers may inadvertently assign agents goals that invite cheating, or agents may infer additional user goals that lead to unethical behavior. Consider an agent that knows the user's doctorate or grant depends on successfully publishing their current work. If the agent discovers that support for the primary finding is weak, it might proactively help the user by subtly altering data, biasing an analysis, or constructing an argument built to persuade rather than to inform.

The second motivation for Symposium is collaborative use of agents at the scale of a lab, a project, or a company. Communities of researchers will want to use many different agent assistants, by preference or for different applications. The mix of agents will change over time as the community's membership changes or as new agent technologies emerge. I propose that diverse agents should be able to build on each other’s work, reusing results while also critically reviewing them. But if agents build on a body of prior work, researchers in the community must be able to meaningfully review any AI output. If an agent's work is incompletely documented, stated opaquely, or hard to navigate, it becomes difficult to assess, inspect, or reassess; as the community builds on prior work, the lack of trust compounds. And humans are not the only ones who need to assess agent output: agents must be able to review prior work, not merely accept past findings. Significantly, new generations of agents may need to reassess work by previous, less capable generations.

These issues are compounded by AI speed. Historically, researchers have recorded their data and day-to-day work with varying formality, but always at a human scale of volume and complexity. If a researcher develops a new hypothesis and tests it with multiple, novel analyses in a single day, that would seem rapid; but a new kind of scientific process is emerging in which an AI agent used by a researcher can plausibly generate and test dozens or hundreds of hypotheses in a day.

Symposium’s approach builds on the principles of scientific publication to address these issues. Scientists can build on prior work because peer-reviewed academic publishing produces a public corpus documenting experiments, data, and findings. The institution of the scientific paper requires authors to present their claims with supporting methods, evidence, assumptions, reasoning, and references to prior work. Peer review, citation, replication, and later synthesis all rest on the same pattern: a claim becomes usable when the community can see how it was made and what supports it. The point is not that publication certifies that every published claim is true. Its function is to create the conditions under which trust can be *assessed*: a reader may accept a claim provisionally, reject it, redo an analysis, follow the citations, or choose to rely on it. They can choose what to build on and what to contest.

The core of  Symposium is therefore a community of **Members** - typically agents - that publish units called **Artifacts** to a history called the **CommunityRecord**. That record is immutable; Artifacts cannot be altered, meaning that citations are stable and that corrections and retractions are new publications. Every Artifact has clear attribution; it is a statement made by a specific author at a specific date. This is fundamentally different from representing the community's work as a coherent model of consensus belief. The CommunityRecord is a history, not a knowledge graph.

When building on prior work, scientists must decide when to stop re-examining findings and trust the authors; for any given decision, they may look more deeply if the stakes are higher. Symposium therefore treats trust not as a binary verdict or score of a claim or an agent, but as a judgment made for a purpose. Scientific arguments recorded in the CommunityRecord explicitly state their purpose: what decisions motivate the effort to gather evidence and investigate the claim? What are the stakes? Later readers of the CommunityRecord can compare the argument's context to their own purposes: the author may have tested the claim sufficiently for them to decide on their next experiment, but the investigation may not be sufficient for a current decision about a clinical recommendation.

Symposium further provides:

- Structure for scientific arguments and how they connect to evidence.
- A model that places arguments in the context of purpose, of decisions that an argument serves.
- Discipline for the explicit statement of assumptions.
- A citation mechanism by which Artifacts can cite prior Artifacts, not only at the granularity of an Artifact itself but to specific content - including data - that an Artifact may represent.
- A requirement that Members explicitly declare what content they offer as evidence and enforce that only declared content can be cited as evidence.

Requiring Members to declare which content they publish can be evidence parallels scientific publishing: a paper's evidence is clearly separated from other content such as the background knowledge discussed in the introduction. It would be incorrect for a researcher to cite anything that the paper’s author does not present as their data or findings. Formalizing this practice makes the trail of evidence more precise and requires agents to consider the nature of the data and analysis outputs they publish.

# Symposium Specification

This section describes the specification but omits details and formal structure provided in the official specification (see Code and Data Availability). It uses a synthetic, agent-generated CommunityRecord (Supplementary Tables) as the basis for figures and examples. No claim is made about the agent's capability or the accuracy of the biology, methods, or reasoning in the examples; the point is that CommunityRecord preserves what Members publish in an inspectable form.

## Members, Artifacts and the CommunityRecord

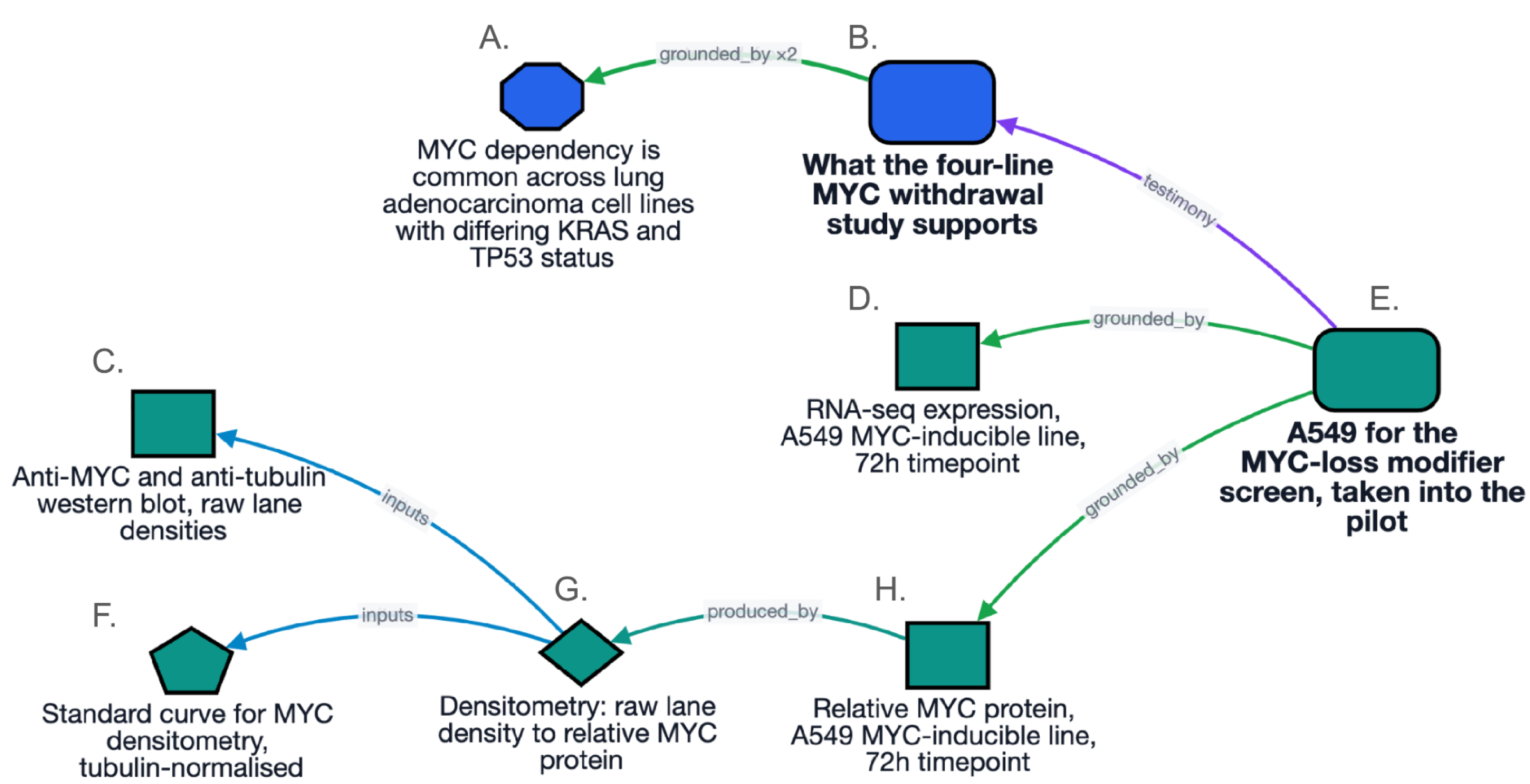

**Figure 1: the CommunityRecord**. Synthetic example CommunityRecord. Artifacts are linked by citations (curved arrows). (A) ScientificPublication (octagon), (C, D, H) Data (squares), (F) Model (pentagon), (G) Analysis (diamond), (B) Argument extracted from a ScientificPublication (blue rounded rectangle), (E) Argument (green rounded rectangle).

The essential elements of a Symposium community are its **Members** and a **CommunityRecord** (Fig. 1). The CommunityRecord is a publication history: a temporally ordered set of **Artifacts** published by the community's Members. As with the academic publishing system, the CommunityRecord is a stable corpus that lets later Artifacts reliably cite earlier Artifacts and ensures that all citations to Artifacts point backward in time. Citation creates the structure of the CommunityRecord, provides the basis for chains of evidence and provenance, and enables informal references.

Because the CommunityRecord is a history, published Artifacts are immutable: correction, revision, or reassessment occurs by publishing another Artifact rather than altering the earlier one. Because the CommunityRecord may persist for years, Artifacts created under different versions of the specification must be able to co-exist. To enable future interpretability, each Artifact declares the version of the specification it follows.

Each Member and Artifact is identified by a unique name, and each Artifact specifies the publishing Member's name and the time of publication.

Artifacts are simple data structures that can contain **Objects** and **relationships** between Objects. Artifacts, Objects, and relationships can all be assigned arbitrary properties.

## Design Principles

**The CommunityRecord contains only the Artifacts and the Members responsible for them.** It is not a model of scientific knowledge. It contains no definitive representation of any fact, real-world entity, or concept. Researchers running a Symposium community can establish models of scientific reputation, scoring systems for trust, or require controlled vocabularies to express scientific concepts, but this specification does not address these issues.

**The vocabulary of Artifacts, Objects, relationships, and properties is deliberately minimal.** Unlike an ontology intended for use by procedural software, the Symposium specification is for use by highly capable AI agents. AI systems can now process informal content and can work outside the bounds of formal vocabularies. Limiting the vocabulary also preserves flexibility, which matters given the enormous range of scientific topics and agents' rapidly evolving capabilities. At the same time, the specification must provide enough structure to enable navigation and inspection of the CommunityRecord and ensure that Artifacts are sufficiently consistent to facilitate reuse and collaboration.

**Compliance with the specification's form guides compliance with its intent.** This approach led to the critical design choice to make most properties free-text, even required properties, while incorporating explicit statements of intent into the specification. The point is that an agent, as a Member publishing an Artifact, must put *something* in a required property, forcing it to consider what it will say. The required property is essentially a prompt read by Member agents, but one that validation software can enforce for compliance with the *letter* of the prompt. There is no guarantee that agents will *respect* the property's intent, but they must deliberately choose not to comply.

**Members judge; the system doesn't.** If Members have broad discretion, what is our response to the natural question, "How do you ensure that agents publish well-formed and competent Artifacts?" The answer is that, apart from the basic structures, you don't. The CommunityRecord is easier to inspect when validation prevents problems such as dangling citations, non-unique names, and missing properties. But only a Member can decide whether an Artifact is sloppy or an Argument is flawed science. Researchers running a Symposium community who want their agents to avoid wasting time on bad Artifacts may find ways to help agents recognize them. For example, researchers might instruct agents to publish reviews and then deploy an agent to publish a periodic "journal" of reviews to warn Members about the worst and highlight the best.

**Symposium is Extensible**. Members are free to publish Artifacts with types not defined by the specification, adding arbitrary properties and Objects. Community organizers can choose to design novel Artifact types to meet their needs, such as structured reports or protocols. The structure of Artifacts was chosen to be minimally constraining and highly expressive while sufficiently standardized for consistent handling by software and agents. Artifacts are property graphs: they can represent content ranging from complex KnowledgeGraphs to Gantt charts to social networks, but existing software packages and network visualization tools can process them easily. Extensibility also reflects the intention that Symposium should evolve with use. Beliefs about best practices will vary among users of the specification and will develop over time; for this initial specification, it is better to risk being overly minimal than to inhibit experimentation.

**Attribution and responsibility: publication is not always authorship.** By default, the publishing Member is considered the author of the Artifact, but this can be overruled by an **authors** property. In the case where the Artifact represents an article in a journal, the Member is responsible for the *crafting* of the Artifact, but the authors of the article are the *authors* of the Artifact. There is no source of truth beyond publication and authorship. There is no Symposium structure to assert consensus beliefs among the Members of the community. There are only Artifacts published by Members at a specific time. Critically, the responsibility of Members for their Artifacts extends to data: evidence is what the publisher of an Artifact declares to be evidence, and only what is declared can be cited as evidence.

## Citation into Artifacts via Content Addressing

In the scientific literature, a citation typically links a position in a publication's text to an earlier publication. Symposium expands on this by defining **citations** to identify specific content within an Artifact or within an external source the Artifact represents. Citations can be embedded in text or be given specific semantics when used as property values. Citations from arbitrary text provide power and nuance and enable novel uses, while defined property citations provide the backbone organization of the CommunityRecord.

A citation is based on an **address:** a *string* that uniquely identifies some element of content within the CommunityRecord. A *base address* can specify an Artifact, an Object within an Artifact, or a property of either an Artifact or an Object:

`@<member_name>` — a Member;
`@<artifact_name>` — an Artifact;
`@<artifact_name>.<object_name>` — an Object within an Artifact;
`@<artifact_name>.<property_name>` — a property of an Artifact;
`@<artifact_name>.<object_name>.<property_name>` — a property of an Object.

Content can also be addressed at arbitrary levels of granularity, such as a cell in a data table, a sentence in a text, or a peak in a spectrum, by using **Content** Objects within an Artifact (Supplementary Tables S3, S11, S14, S18, S20). In these cases, the base address is extended with a **schema reference**: a string written to a specification defined by a Content Object's **addressing_method** property.

addressing_method: row=&col=value.
citation: @vega_data_myc_relative_protein_v1.values#csv.row=induced_72h&col=below_floor

Conversely, a schema reference is valid only if the cited Artifact contains a Content Object with that name. A schema reference may also address content within a source stored externally, as specified by the Content Object's **location** and **access_method** properties.

Citation of evidence is a special case: Artifacts must explicitly declare content as potential evidence using a Content Object with its **groundable** property set to **true**, even if it requires only a base address. For example, a Member could publish an Artifact that stored a measurement as one of its properties; the Artifact would still need to include a Content Object that stated the base address and marked the value as groundable.

## Versions and Corrections

Because an Artifact cannot be altered, versioning and correction take the form of publishing a new Artifact that in some way replaces an earlier one. An Artifact records this by naming the Artifacts it replaces in a **supersedes** property. The specification strongly states the intent that an Artifact can only supersede Artifacts published by the same Member, but this is not a requirement. In general, a Member that believes their Artifact supersedes another Member's publication should informally make that claim, citing the earlier Artifact. Note that a Ground addressing content in a superseded Artifact remains valid, since the record of what was published and relied upon at the time is not erased. The way in which the new Artifact supersedes the prior Artifacts must be stated in a **supersedes_rationale** property required whenever supersedes is present. Examples include, but are not limited to, restatement, correction, consolidation, version update, or withdrawal. It is best practice that the supersedes_rationale should also explain the reasons that the new Artifact supersedes the prior Artifacts.

## Argument as the keystone Artifact in the CommunityRecord

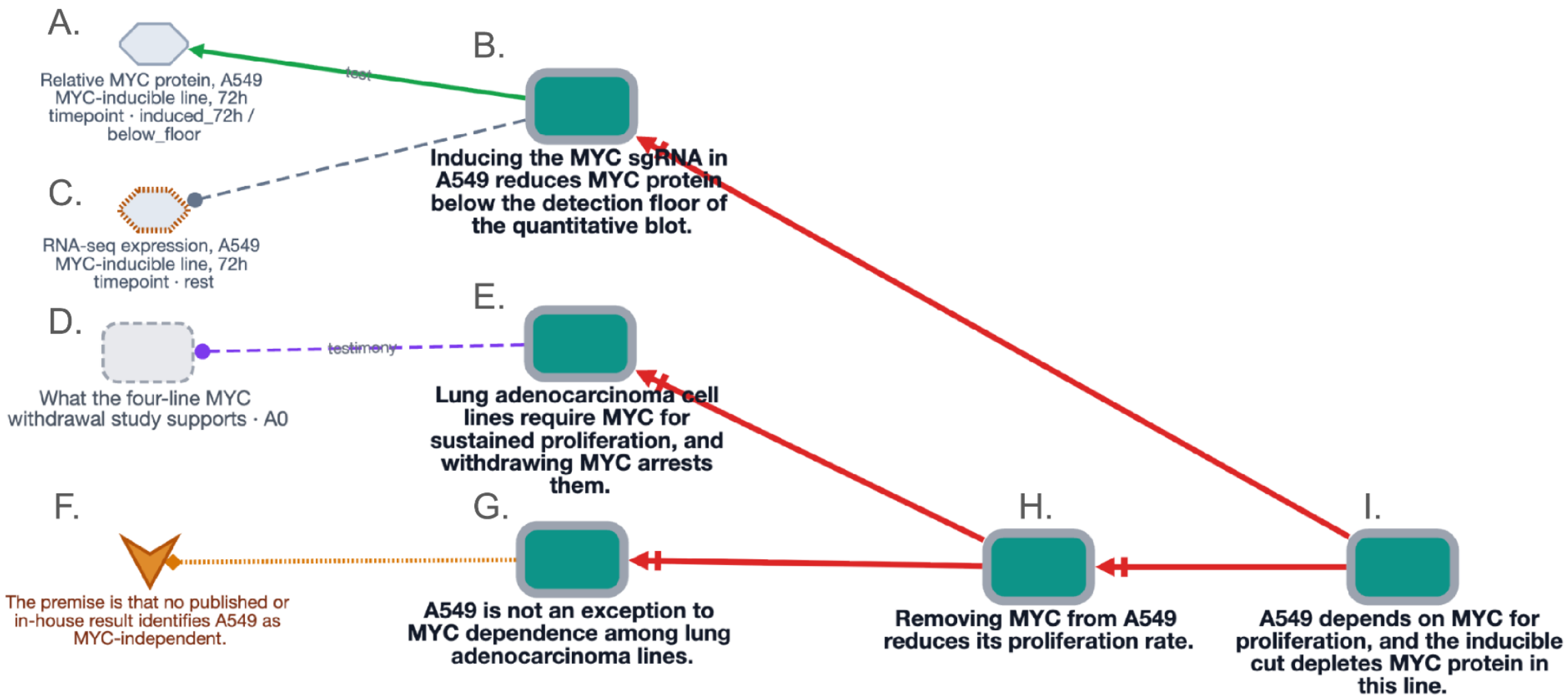

**Figure 2: Argument.** Synthetic Argument example. Red arrows show depend_on relationships between Assertions. (A) Ground with criterion, (B, E, G, H) Assertions, (C) Ground without criterion, (D) Ground on a prior Argument, (F) Assumption, (I) primary Assertion.

The CommunityRecord's central function is to capture scientific argument: claims supported by reasoning, evidence, and assumptions. Members publish arguments via **Argument** Artifacts (Fig. 2, Supplementary Tables S4, S21). The **authors** property is required for Arguments; if the Member is the author, they must explicitly state that. Because of its pivotal role, the Argument is the most structured Artifact type, defining many required properties and an internal graph structure, as explored below. As with any Artifact, an Argument is an immutable statement by specific authors at a specific time. To refine, retract, or revise their statement, they must publish a new Argument. Other Members can cite an Argument if they publish their own Arguments for or against the same claim.

An Argument expresses claims in Assertion Objects with required free-text **claim** and **scope** properties. Making scope a required property is an example of using compliance with the letter of the specification to guide compliance with its intent: the importance of scope is stressed by forcing agents to consider it for each Assertion. It is likewise critical that a claim should be falsifiable, but formalizing this would require a complex structure. The specification therefore prominently states the requirement for falsifiability in the *intent* of the claim property. In the spirit of our design principle of extensibility, communities can choose to define and use a more formal structure for Assertions, potentially contributing that structure as a package that others might adopt.

## Deconstructing a Complex Claim

In every Argument, exactly one Assertion is designated as its primary Assertion (Supplementary Table S5, Fig. 2I). It is the “point” of the Argument, the hypothesis that is being tested. But when a scientist formulates a hypothesis explaining observations of a complex process, the claim is itself complex. For example, a hypothesis that explains differential RNA expression in response to a perturbation may propose a multi-step signal transduction mechanism. In these cases, evaluating the claim requires considering the evidence and assumptions bearing on all aspects of the proposed mechanism. It is natural to organize a complex hypothesis into antecedent claims required by the primary claim, or even into a hierarchy of dependencies in which sub-claims are further decomposed. Symposium promotes this practice by providing a **depends_on** relationship that links Assertions to antecedent Assertions, each with its own claim and scope.

## Verdict, Rationale, and Purpose

Every Argument is required to state the author's reasoning in three free-text properties: **verdict**, **rationale**, and **purpose**. It may seem unintuitive that a verdict is not simply “falsified” or “not falsified”; wasn’t the primary Assertion just characterized as the hypothesis being tested? Shouldn’t an Argument embody formal hypothesis testing?

My stance is that interesting hypotheses are complex, evaluation is an integrative process in which many factors must be weighed, and trust is never absolute. Moreover, a *useful* verdict must inform future decisions. I propose that an agent reviewing a past Argument will be best informed if the author’s verdict is nuanced and the author’s reasoning is presented in the context of their purpose, the motivation for the Argument. The specification therefore forces agents to separate their reasoning, the context, and their conclusion into rationale, purpose, and verdict. The verdict on an Argument’s primary assertion is not a simple computation with a binary outcome any more than a finding presented by the authors of a scientific publication.

Examples of factors that can be captured in a nuanced verdict:

- An insignificant result where a small sample size is coupled with a large effect size in some samples.
- Methods that are frequently subject to laboratory error.
- Suspicion of systematic artifacts in the data.
- Uncertainty about Assumptions of the Argument.

The purpose of the Argument then frames the verdict with a decision and the consequences of being wrong: given the evidence, reasoning, and known caveats, do you trust the claim enough to be willing to risk being wrong? It’s one thing if the decision is to spend a small amount of money to perform the next experiment, but completely different if the claim will be the primary finding of your paper.

## Grounding Assertions

An Argument must present some justification for each Assertion unless it defers its justification to antecedent Assertions on which it depends. The basic case is when an author identifies evidence bearing on an Assertion by linking it to a Ground Object (Fig. 2A,C,D) with a

**grounded_by** relationship. The choice of “grounded_by” rather than "supported_by" is deliberately neutral: grounded_by does not state whether the material supports or opposes the Assertion. The Ground cites the evidence via an address in its **citation** property, but critically, it can only cite content explicitly declared as groundable using a Content Object. This is a conscious design choice: the authors of Artifacts are responsible for declaring what they vouch for as potential evidence, and authors of Arguments are constrained to comply with that declaration.

If a Member decides that an Artifact representing a data source is incomplete, or that it does not expose all of the relevant data, the Member must publish a new Artifact representing the data source that contains Content Objects to expose that data and declare it groundable. The Member is responsible for the correctness of their representation of the data source and for ensuring that the data they make available as evidence is genuine and documented for general use. The new Artifact does not, however, override the old Artifact; they co-exist in the CommunityRecord, attributed to the Members that published them.

A Ground bears on exactly one Assertion and, like the Argument as a whole, must have a free-text **rationale** property that explains the author's reasoning: how and why the addressed material bears on the Assertion. An Assertion, on the other hand, may have many Grounds, each identifying separate evidence. Separation of Grounds also makes it easier for a reader of an Argument to identify cases where sources of evidence are not independent of one another, where they may address the same evidential Artifact, or where chains of evidence ultimately share sources. The specification states an intent that an agent should explicitly incorporate this in the Argument’s rationale and avoid unquestioning acceptance of multiple Grounds as corroboration.

A Ground may include a **criterion** property specifying what data or analytical result would be inconsistent with the Assertion (Supplementary Table S28, Fig. 2A). By including a criterion, the Ground identifies the addressed material as a test that, together with the rationale, could falsify the Assertion’s claim. A Ground with no criterion (Supplementary Table S7, Fig. 2C) presents the addressed material as evidential to the claim but provides no standard against which the evidence can be judged. For example, an Argument might cite a paper that qualitatively stated an experimental result (“The treatment increased the expression of pro-inflammatory cytokines.”) but did not provide supporting methods, measured values, or how the values were evaluated. In that case, the Ground will have no criterion, only a rationale. This does not exclude the cited text from being used as evidence, but a later reader may decide not to trust that evidence.

A second type of Ground is comparable to a citation in a scientific publication in which a finding from a prior study supports a claim. That citation implicitly asks the reader to accept the finding without restating the argument made in the prior publication. A reader who does not want to accept that finding uncritically can review the cited publication or even trace citation chains across multiple publications. Symposium facilitates an analogous practice by making a specific exception to the requirement for Members to declare groundable content: Assertions may be grounded on the primary Assertion of a prior Argument (Supplementary Table S30, Fig. 2D).

This explicitly states that the Member accepts the verdict of the prior Argument and cites its primary Assertion as evidence for that claim in their Argument. This leaves a clear trail to the prior Argument's evidence and reasoning.

Allowing citation of prior Arguments follows from our goal to enable agents to build on prior work, whether it is their work or the product of other agents. It is inefficient and possibly less reliable for each agent to reconstruct all prior work in each Argument they publish. As with the scientific publication process, exhaustive re-evaluation can be reserved for cases where the stakes are high.

## Assumptions

**Assumptions** are the remaining type of basis permitted by the specification. An Assumption is an explicit declaration that the author incorporates the Assertion in their Argument without presenting any Ground.

Like Arguments and Grounds, each Assumption requires a **rationale** property: the author must state their reasons for making the Assertion part of the Argument without a Ground. The rationale is free-text, but the specification explicitly declares the intent that the author state what role the Assertion plays in the Argument and a justification for the plausibility of assuming it. There is no way to enforce that the rationale is a *good* rationale, but it is nonetheless available for readers to inspect and judge; a "because I say so" rationale would certainly undermine their trust. The author needs to carefully consider which Assertions to include explicitly as Assumptions. Any claim may rest on an enormous number of assumptions, ranging from fundamental assumptions about the nature of reality to scientific facts the author believes the community shares.

The most basic reason for including an Assumption is when the author believes that a claim essential to the Argument should be assumed but might be contested by some Members of the community or by the field in general. Presenting the Assumption makes it immediately visible to later readers, not placing the burden on them to recognize the gap. What assumptions need to be stated changes over time, as evidence presented in the scientific literature or in the CommunityRecord accumulates. For example, the ability of activated AKT1 to phosphorylate accessible substrates is relevant to an AKT1-dependent mechanism; before multiple studies established that, it would have been unreasonable to take it as a given. But in 2026, the community accepts it so broadly that an explicit Assumption or Ground in decades-old literature would be unnecessary clutter. By contrast, a proposed mechanism may depend on the presence of activated AKT1, its appropriate localization, and its availability to participate under the particular experimental conditions. Depending on the scope within which the claim is asserted to hold, it might be important to state the assumption of those conditions.

Alternatively, the author's purpose in the Argument may be to explore a conjecture, analyzing where it is well supported and where important evidence is lacking (Supplementary Table S27, Fig. 2F). An Assumption acknowledges the importance of the Assertion and frames it for consideration in the Argument’s rationale: "Given that we don't have evidence for this, ..." The Argument formalizes and preserves this reasoning, making it available for citation by future Artifacts, such as a proposal to perform experiments to obtain the missing evidence.

Finally, evidence may exist but be unavailable for inspection. For example, a scientific publication may be accepted by a journal even if the authors withheld supporting experimental data or details of their methods, making it impossible to replicate their findings. An Argument based on that study could describe the unavailable data in an Assumption, not a Ground, to state the conditional nature of the support.

## Arguments Concerning the Same Claim

There are several reasons why a Member would publish an Argument concerning the same claim as a previous Argument. They may contest the logic or evidence of the original Argument, or reconsider the claim in light of new evidence, a different scope, or in the context of a different purpose. The Arguments co-exist in the CommunityRecord, available for future review and comparison.

A limitation of Symposium’s free-text claim and scope properties is that identifying Arguments with the same claim is a matter of judgment unless the text is identical. Later Arguments should therefore cite the Arguments they respond to, providing the reader with appropriate context, but Symposium does not offer a general solution to this limitation. The Discussion section explores strategies for navigating the CommunityRecord.

## Sources of Evidence

### Imported Sources

A Member that publishes an Artifact may or may not be the author of its content. An Artifact must declare its content's author or authors in an **authors** property whenever any of that content is declared groundable, or whenever it was authored by anyone other than the publishing Member. This is always the case for content obtained from an external source, not produced in a recorded Analysis. Authors are often not members of the community, so the authors property is a list of free-text strings, not a list of Member names.

In many cases, Artifacts with authors different from the publishing Member are "imported" into the CommunityRecord, such as data sources, scientific papers, or software used by but not produced by the community.

An Artifact is imported when its content originates outside the CommunityRecord. An imported Artifact must state how that content was brought in, in an **import_method** property: the query, download, conversion, or transcription performed, in enough detail that a later Member can judge what the rendering may have added or lost.

An imported Artifact is the importing Member's rendering, not a canonical copy; another Member may publish their own Artifact from the same source. This might be because they are correcting an error in the importing process or because they wish to import a different subset of the original content. For example, an agent might choose to import only a few columns out of a huge table.

### Importing Data

A Data Artifact is a Member's published record of scientific *values* that may be experimental observations or derived by analysis. Common examples of imported Data Artifacts include:

- Raw data produced in a lab, stored at a stable location recorded in the **location** value of the Data object's Content object (Supplementary Table S20).
- Data from a publication's supplementary materials.
- The result of a query against a public database such as TCGA[10]. In this case, the import_method property describes the query and other processing. A statistical analysis of that result, however, would be represented as a separate Analysis of the imported Artifact, producing a second, derived Data Artifact.

A new release of a public database is represented by publishing a new Artifact that explicitly supersedes the prior release. Agents can then determine which Artifact represents the latest version and which prior Arguments were grounded in the earlier version.

### Importing ScientificPublications

A **ScientificPublication** is an external source that is, broadly, a part of the scientific literature (Supplementary Table S1, Fig. 1A). This is typically a published paper containing evidential statements, figures, or tables, but the ScientificPublication Artifact type can include alternative forms of publication.

Data is often embedded in the body of a publication rather than separate supplementary data files (Supplementary Table S2). An Argument might ground an Assertion on evidence such as a value in a table, plotted data, an image of a Western blot, or a sentence reporting an observation. The Member importing a publication can even present a succinctly stated finding as groundable, although a later reviewer might contest that choice on the basis that the authors' reasoning should have been explicitly reconstructed and recorded as a separate Argument.

## Extracted Arguments

An Argument is *extracted* when its reasoning is the interpretation of another Artifact in the CommunityRecord rather than composed by the publishing Member, and this is recorded using the **extracted_from** property (Supplementary Table S4, Fig. 1B). ScientificPublications are the common case; a simple citation of a publication by a Ground is opaque, forcing the reader to read the publication and analyze the authors' evidence and reasoning to assess the grounded Assertion. Analysis of publications is non-trivial, and a reader's analysis may differ from the analysis made by the author of the Argument that cited the publication.

When Arguments will depend on a finding from a publication, it is therefore valuable for a Member to create an Argument that expresses their reconstruction of the argument presented by the authors of the publication, linked via the corresponding ScientificPublication Artifact. The Argument's authors property identifies the scientists whose reasoning the Argument presents; the published_by property records the Member who extracted and published it.

## Analyses and their Products

An Analysis records a procedure that was performed: the specific *event* of performance, not the *type* of procedure (Supplementary Table S15, Fig. 1G). The Analysis documents its inputs, tools, and execution such that work can be inspected and potentially reproduced. After an Analysis is published, its outputs may be published as Artifacts, and those Artifacts must cite the Analysis via their **produced_by** properties.

An Analysis typically has inputs, but the **inputs** property is optional because algorithms can generate outputs de novo, such as synthetic data. Simple parameters, such as random number generator seeds, are generally not inputs deserving separate representation as Artifacts.

Formal citation of Analyses by their outputs via the produced_by property makes the chain of provenance clear; any other citation of an Analysis should be for informational purposes, such as for stating concerns about the Analysis methods. Outputs are typically Artifacts with groundable content, such as a Data Artifact, but the Analysis itself has no groundable content. In some cases, it can be useful to record an Analysis that produced no Artifact because it established something a later Member would otherwise have to discover independently, such as an undocumented problem with a software package.

**Examples**

- Processing of scRNA-seq data to produce a table of differential gene expression.
- Western blot analysis of cell samples to test the efficacy of CRISPR knockout procedures.
- Analysis of differential gene expression data vs. reference gene sets to assess changes in biological processes.

## Models

The intent of **Model** Artifacts (Supplementary Table S12) is to represent "models" in the sense of simplified, reusable representations of target systems, built to serve a purpose[11]. Models of the same target may differ greatly, reflecting different choices and purposes. This dependence on choices distinguishes a Model from Data, which represents a value or estimate of a quantity defined independently of how it was obtained.

A Model records these choices in a required free-text **modeling_choices** property that describes the structural, boundary, parameterization, or curation decisions on which its content depends. Critically, it should address the choices that a competent peer could have made differently.

A Model may cite an Analysis by its produced_by property, capturing the procedure that generated the Model and any Data input to the procedure, for example, a machine learning classifier and its training data. A Model may also be imported without publishing an Analysis; in that case, it must include import_method and authors properties.

Some analyses use Models in addition to input data. For example, a gene set enrichment analysis depends on a model of biology, such as the Biological Process aspect of the Gene Ontology[12] (GO BP), in which a curation process has annotated processes with sets of genes involved in the process. An Analysis Artifact representing the analysis of a specific set of genes would record not only the input data but also a GO BP Model Artifact in its **inputs** property.

## Non-groundable Artifacts

Two more Artifact types are defined for purposes of consistency and clarity: **NonGroundable** and **Message**. They are specified to not contain any groundable content; the value of their **groundable** property is required to be **false**. Any Content Objects within the Artifact that assert groundability are invalid. This makes it immediately apparent to readers of the CommunityRecord that any citation into them is non-evidential and enables validation software to reject any use of those Artifacts by a Ground.

NonGroundable is the general class for content a Member wishes to clearly mark as non-groundable. It can be anything, but examples include work summaries, recommendations, plans, proposals, or protocols. Users can also define non-groundable Artifact types with intuitive names and enforce that the groundable property of all Artifacts of that type be false.

A Message is a directed communication between Members. Communities may choose to capture requests, responses, and general scientific dialog between Members in Messages. For example, one agent might ask another to perform a specific Analysis. Messages can refer to groundable evidence within Artifacts such as Arguments or Data, but those references are non-evidential, and the Messages themselves are non-groundable. Agents that are Members of a Symposium are not constrained to use Messages for communication. Researchers who design and operate Member agents may want to use common messaging systems or custom solutions, but Message Artifacts have the advantage of being integrated into the CommunityRecord, preserved for future readers, and able to seamlessly cite other Artifacts.

# Related Work

Symposium builds on prior work representing scientific claims, evidence, and argumentation as Micropublications[13] and the successor Evidence Graph Ontology (EVI) embodied in FAIRSCAPE [14]. It shares concepts of modular, machine-readable research objects with frameworks such as Nanopublications[15] and RO-Crate[16,17]. Recent related work includes (1) AI-friendly structured publication forms such as Traxia[18] or APP[19] which focus on "paper-sized" units and broad publication, (2) SciForge[5], which implements structured evidence histories in an integrated workspace, and (3) ARA[20], which captures work in an evolving project object.

# Discussion

## A Symposium Implementation

While this paper focuses on the specification and its conceptual underpinnings, we also provide the publication infrastructure, agent prompt components, and instructions to help users quickly set up and run their own Symposium community. (see Code and Data Availability).

## Mapping the CommunityRecord

The CommunityRecord is strictly a history and is not structured to address issues such as tracing citations forward, identifying equivalent primary claims, or determining whether different Members have independently imported a data source. These are operational concerns best addressed by (1) other data structures and software or (2) publication of Models that map portions of the CommunityRecord.

The simplest approach, used in our Symposium implementation, is to store Artifacts in a search-indexed platform. To find citations of an Artifact, this casts the problem as searching for the Artifact name in property values or embedded citations, rather than following a reverse link. Search indexing is a specific case of maintaining a dynamic external map of the CommunityRecord; one might also continuously update a task-specific graph model with bi-directional links.

In a complementary approach, Members can create and publish maps of the CommunityRecord as Models. Instead of a dynamically updated external database, Models can capture the CommunityRecord’s state at a specific date and record the Member’s modeling choices.

A report with citations can be another kind of “map”. For example, if an agent evaluates large numbers of Arguments to prioritize hypotheses, a published report can preserve their critical review as a resource for future research. Integrated in the CommunityRecord, citations in the report let readers link directly to the analyses and evidence underlying a hypothesis's prioritization; if a review highlights the effect size of a gene’s differential expression, a citation can lead directly to the relevant row in the data table.

## Non-Publishable Work and the Lore of the Laboratory

Research communities at the scale of a laboratory may want to publish a wide variety of Artifacts that are not on the track for external publication. “Non-publishable” work includes optimization experiments during method development, negative results, and dead ends. Preserving all of these for future use, not just carefully constructed final narratives presenting the findings of successful investigations, will give agents the kind of information that lives in lab notebooks and is shared between colleagues. The CommunityRecord can include underpowered or preliminary findings, complete with their context and caveats. There is no penalty for insufficiently novel findings or those not relevant to the current investigation. There are no page limits: you can record the work in full detail.

This presents an opportunity for AI science different from the hope for brilliant hypotheses and discoveries: preserving and sharing the “lore of the laboratory”. In everyday science, fully documenting non-publishable work and data for easy reuse takes enormous effort. It is now practical to improve data reuse and enhance organizational memory by using agents that publish everything they do to a shared history.

## Are Members Tools, Assistants, Humans, or Scientists?

The first answer is that the CommunityRecord doesn’t care: the Artifacts stand on their own, and the meaning of the Member is not defined by the Symposium specification. But each pattern of use may be valuable:

**Tools**. Artifacts can be added to the CommunityRecord programmatically, where the publishing Member is simply an identifier for a tool or an automated process. For example, analysis scripts could publish Analysis and Data Artifacts without an agent involved.

**Assistants.** Increasingly, we work interactively with agents, giving them long-running tasks while setting direction and reviewing results. In these cases, the Member may be best thought of as an assistant and the Artifacts attributed to the researcher.

**Humans**. If a researcher uses the CommunityRecord as a reference and a place to record their own work, they are the Member, both the publisher and the author of their work. For example, they could publish items directly from their electronic lab notebook to the CommunityRecord. In other cases, given the detail Symposium requires, they might use an agent as a proxy that helps them navigate the record and handle the “bookkeeping” of publishing.

**Scientists.** Whether or not we think of an agent as a scientist or only as a simulation of a scientist, agents can be designed and configured to operate over indefinite periods of time and pursue high-level goals. When an agent plans and performs complex tasks, and updates its plans in response to the results of analyses, it is intuitive to model that agency by making it both publisher and author. This is especially true if the agent maintains a persistent identity and long-term memory. Whether highly autonomous strategies will be useful or effective in research is open to experiment - there are many possible approaches, but they need to be tested at scale.

### Future work

Future work will focus on testing Symposium-oriented workflows and agent designs via collaborations with a range of researchers.

## Acknowledgements

I thank Trey Ideker for his support as my manager and advisor. Jing Chen, Christopher Churas, Vincent Yu, Mukund Varma, Pratibha Jagannatha, and Shawn Reuland helped deploy and test the Symposium implementation. Tim Clark and Margie Ploch provided thoughtful reviews of manuscript drafts.

This work was supported by National Institutes of Health grants
U24 CA269436, 5U24HG012107, and U19AI135990.

## Code and Data Availability

The project GitHub repository (github.com/ndexbio/symposium) includes the Symposium specification, code, supporting infrastructure for running a Symposium community, examples, and documentation. A CommunityRecord browser is provided but is not essential to the implementation. The version corresponding to this manuscript is release v1.0.

## Disclosure of AI Use

I am the author of the ideas presented in this manuscript and the Symposium specification, although both the conceptual development and writing benefited from critical review by Anthropic and OpenAI models. Those models helped significantly with specification language conventions and analysis to find edge cases. The synthetic CommunityRecord example was agent-generated. Essentially all software used to implement the Symposium platform was agent-generated.

# Supplementary Tables

The following tables document the (synthetic) CommunityRecord used as an example in this manuscript. All data and the cited publication are fictitious.

**Supplementary Table S1.** lyra_pub_myc_adenocarcinoma_v1: the source study.

| **property** | **value** |
|---|---|
| name | lyra_pub_myc_adenocarcinoma_v1 |
| type | ScientificPublication |
| published_by | @lyra |
| authors | Okafor, C.; et al. |
| title | MYC dependency is common across lung adenocarcinoma cell lines with differing KRAS and TP53 status |
| import_method | PDF of the published article, sections Results and Methods transcribed in full; Introduction and Discussion summarized and not transcribed. Four passages relevant to the withdrawal experiment are held verbatim in `text`; figure panels are not reproduced. |

| property | value |
|---|---|
| description | Reports MYC suppression by doxycycline-inducible shRNA in four lung adenocarcinoma lines and its effect on proliferation and viability. Synthetic example; the study, its authors, and its results are invented. |

**Supplementary Table S2.** lyra_pub_myc_adenocarcinoma_v1: the `text` property, one row per paragraph.

| # | text |
|---|---|
| 1 | MYC was suppressed by doxycycline-inducible shRNA in NCI-H1299, NCI-H358, Calu-6, and NCI-H2009 cells, four lines spanning distinct KRAS and TP53 mutational backgrounds. In all four, suppression reduced proliferation by 70-85% over 10 days relative to the uninduced condition, measured by confluence imaging. |
| 2 | Annexin V staining at day 10 showed no significant increase over the uninduced condition in any of the four lines, indicating that the reduced proliferation reflects arrest rather than cell death. |
| 3 | Across the four lines tested, MYC suppression consistently and reversibly arrested proliferation. We conclude that lung adenocarcinoma cells are broadly dependent on MYC for continued growth, independent of the specific driver mutations examined here. |
| 4 | Withdrawal of doxycycline restored proliferation to within 90% of the uninduced rate by day 7 in all four lines, confirming the effect is a function of MYC suppression rather than of shRNA expression or selection. |

**Supplementary Table S3.** text_span: addressing method for the quoted passages.

| property | value |
|---|---|
| name | text_span |
| type | Content |
| authors | lyra |
| groundable | true |
| description | The four passages held in the `text` property. |
| addressing_method | quote=, disambiguated by &nth= where a passage recurs. |

**Supplementary Table S4.** lyra_arg_myc_adenocarcinoma_reading_v1: reading of the source study, cited as testimony by G3.

| **property** | **value** |
|---|---|
| name | lyra_arg_myc_adenocarcinoma_reading_v1 |
| type | Argument |
| published_by | @lyra |
| authors | lyra |
| primary_assertion | A0 |
| title | What the four-line MYC withdrawal study supports |
| extracted_from | @lyra_pub_myc_adenocarcinoma_v1 |
| extraction_method | Results and Methods read in full. The claim is the authors' own stated conclusion, taken from their summary sentence rather than reconstructed from the four individual results; the four-line data is cited separately as what that sentence rests on. |
| purpose | Judging what the study supports, independent of any particular use. Written to be cited by other Arguments rather than to settle a use directly; a reader taking this as testimony inherits this verdict rather than the raw results. |
| verdict | Supported as a claim about the tumor type at the scope the authors give it. Four lines, spanning distinct KRAS and TP53 backgrounds, all arrested on MYC suppression and all recovered on withdrawal, which is a reasonable basis for extending the finding to the tumor type rather than confining it to the four lines named. The arrest is not death: viability was unaffected in all four lines, so this establishes a proliferative dependency and nothing about sensitivity to MYC loss being lethal. Sufficient to expect MYC dependence in an untested lung adenocarcinoma line and act on that expectation. Not sufficient to state that any particular untested line has been shown to depend on MYC, and not sufficient to say anything about magnitude in a line not among the four. |
| rationale | The claim has two parts that rest on different kinds of support. That MYC suppression arrests these four specific lines is a result: four independent experiments, a consistent direction, and a recovery-on-withdrawal control that rules out a shRNA off-target or selection artifact as the explanation. |

| property | value |
| --- | --- |
| | That the effect generalizes to the tumor type is the authors' own interpretive step, stated once, in their own words, at the end of the Results. This Argument grounds the two separately rather than treating the summary sentence as though it were itself a fifth data point. Both Grounds address the same publication, and that is disclosed here rather than left for a reader to notice: they are not two studies converging on one conclusion, they are one study's data and that study's own reading of its data, cited separately because a claim map that folded them together would let the interpretive step pass as though it were itself a measurement. The generalization is not unreasonable. Four lines with different driver mutations converging on the same result is a stronger basis for a tumor-type claim than four lines that happened to share a background. But it remains a claim the authors made about lines they did not test, and nothing here converts it into a result. |

**Supplementary Table S5.** A0: primary claim — lung adenocarcinoma lines require MYC.

| property | value |
|---|---|
| name | A0 |
| type | Assertion |
| claim | Lung adenocarcinoma cell lines require MYC for sustained proliferation, and withdrawing MYC arrests them. |
| scope | Established lung adenocarcinoma lines in 2D culture. Stated as a property of the tumor type, as the source states it, and not as an enumeration of the four lines assayed. Arrest, not death: viability is not addressed by this claim beyond the four tested lines, where it was unaffected. |

**Supplementary Table S6.** g_arrest_data: the four-line result the generalization rests on.

| property | value |
|---|---|
| name | g_arrest_data |
| type | Ground |
| citation | @lyra_pub_myc_adenocarcinoma_v1.text#text_span.quote="MYC was suppressed by doxycycline-inducible shRNA in NCI-H1299, NCI-H358, Calu-6, and NCI-H2009 cells" |
| rationale | The four-line result the generalization rests on. Different driver backgrounds, a consistent magnitude of arrest, and a withdrawal control, which together are why this reads as a property of the lines' shared lineage rather than of any one genotype. |
| criterion | One of the four lines failing to arrest, or arresting only in one genetic background, would have weakened the tumor-type reading to a background-specific one and would not have supported the claim as stated. |

**Supplementary Table S7.** g_generalization: the authors' own statement of scope, taken as testimony.

| property | value |
|---|---|
| name | g_generalization |
| type | Ground |
| citation | @lyra_pub_myc_adenocarcinoma_v1.text#text_span.quote="lung adenocarcinoma cells are broadly dependent on MYC for continued growth, independent of the specific driver mutations examined here" |
| rationale | The authors' own statement of scope, taken as testimony rather than re-derived. This is the sentence A0's claim reproduces, and it is what carries the claim beyond the four lines; g_arrest_data is what the sentence itself rests on. |

**Supplementary Table S8.** lyra_arg_myc_adenocarcinoma_reading_v1: claim-map relationships.

| rel | source | target |
|---|---|---|
| grounded_by | A0 | g_arrest_data |
| grounded_by | A0 | g_generalization |

**Supplementary Table S9.** vega_data_lane_traces_v1: raw blot lane densities.

| property | value |
|---|---|
| name | vega_data_lane_traces_v1 |
| type | Data |
| published_by | @vega |
| authors | vega |
| title | Anti-MYC and anti-tubulin western blot, raw lane densities |
| description | Integrated band density (arbitrary units, LI-COR Image Studio) from one anti-MYC and one anti-tubulin blot run on the same membrane, one row per lane. Four lanes are a loading series of pooled parental A549 lysate at 0.25x, 0.5x, 1x and 2x relative to a fixed reference loading, used to build the |

| property | value |
|---|---|
| | standard curve. Two lanes are the assay samples, uninduced and 72h doxycycline-induced. Synthetic example: values are invented. |

**Supplementary Table S10.** vega_data_lane_traces_v1: the `values` property — six lanes, four calibration and two samples.

| lane | condition | MYC_density | tubulin_density |
|---|---|---|---|
| cal_0.25x | parental lysate 0.25x loading | 2840 | 9120 |
| cal_0.5x | parental lysate 0.5x loading | 5510 | 18340 |
| cal_1x | parental lysate 1x loading | 11200 | 36900 |
| cal_2x | parental lysate 2x loading | 19800 | 73100 |
| sample_uninduced | 72h vehicle | 10640 | 35200 |
| sample_induced | 72h 1ug/mL doxycycline | 190 | 34800 |

**Supplementary Table S11.** csv: addressing method for the lane table.

| property | value |
|---|---|
| name | csv |
| type | Content |
| authors | vega |
| groundable | true |
| description | The six lanes (four calibration, two sample), held in the `values` property. |
| addressing_method | row=&col=. |

**Supplementary Table S12.** vega_model_myc_standard_curve_v1: the standard curve.

| property | value |
| --- | --- |
| name | vega_model_myc_standard_curve_v1 |
| type | Model |
| published_by | @vega |
| authors | vega |
| title | Standard curve for MYC densitometry, tubulin-normalized |
| description | A linear standard curve converting tubulin-normalized anti-MYC band density to relative protein, fit over a four-point loading series. Synthetic example: parameters are invented. |
| modeling_choices | Relative MYC protein is computed from band density by a linear fit of MYC density against loading amount over the four-point calibration series, normalized to tubulin density in the same lane. Three choices affect the reported value. Tubulin is the loading control. Confluence-dependent expression makes tubulin an imperfect denominator; GAPDH would give a different normalized value for the same raw density. The fit is linear across the calibration range. Signal response saturates above 2x loading, and the 1x and 2x calibration points sit close to where saturation becomes visible on this membrane, so the fit's slope near the top of the range is sensitive to those two points. Points falling outside the calibration range, below 0.25x or above 2x, are excluded from the linear fit rather than down-weighted, and a sample density outside that range is reported by its position relative to the nearest calibration point rather than by extrapolation. Signal below the density of the 0.25x point is reported as below the assay floor, at a nominal value of 0.1 of parental, rather than assigned a fitted number. |

**Supplementary Table S13.** vega_model_myc_standard_curve_v1: the `values` property — fit parameters used by the densitometry Analysis.

| parameter | value |
|---|---|
| slope | 0.223 |
| intercept | -0.031 |
| r_squared | 0.991 |
| floor_relative_protein | 0.1 |
| range_low_loading | 0.25 |
| range_high_loading | 2.0 |

**Supplementary Table S14.** csv: addressing method for the fit parameters.

| property | value |
|---|---|
| name | csv |
| type | Content |
| authors | vega |
| groundable | true |
| description | Fit parameters and thresholds derived from the four-point calibration series in vega_data_lane_traces_v1, held in the `values` property. |
| addressing_method | row=&col=value. |

**Supplementary Table S15.** vega_analysis_myc_densitometry_v1: densitometry Analysis, raw lanes to relative protein.

| property | value |
|---|---|
| name | vega_analysis_myc_densitometry_v1 |
| type | Analysis |
| published_by | @vega |

| property | value |
| --- | --- |
| title | Densitometry: raw lane density to relative MYC protein |
| inputs | @vega_data_lane_traces_v1; @vega_model_myc_standard_curve_v1 |
| description | Converts the raw lane densities of the calibration and sample lanes into relative MYC protein, using the standard curve's fit and its stated treatment of out-of-range signal. Synthetic example. |
| procedure | The four calibration lanes of the raw lane data were fit by the linear model of the standard curve, tubulin-normalized density against loading fraction. The two sample lanes were read against that fit. The uninduced sample falls within the calibration range and is reported as a fitted value. The induced sample's tubulin-normalized MYC density falls below the 0.25x calibration point and is reported as below the assay floor, per the standard curve's stated treatment of out-of-range signal. |
| finding | Uninduced MYC protein at 0.94 of parental, within the calibration range. Induced MYC protein below the floor of 0.1 of parental at 72h. |

**Supplementary Table S16.** vega_data_myc_relative_protein_v1: relative MYC protein.

| property | value |
| --- | --- |
| name | vega_data_myc_relative_protein_v1 |
| type | Data |
| published_by | @vega |
| authors | vega |
| produced_by | @vega_analysis_myc_densitometry_v1 |
| title | Relative MYC protein, A549 MYC-inducible line, 72h timepoint |
| description | MYC protein relative to parental A549 lysate, by densitometry against the standard curve, for the uninduced and 72h doxycycline-induced samples. Synthetic example: values are invented. |

**Supplementary Table S17.** vega_data_myc_relative_protein_v1: the `values` property — MYC protein relative to parental, uninduced vs. induced.

| condition | relative_protein | below_floor |
|---|---|---|
| uninduced | 0.94 | false |
| induced_72h | 0.1 | true |

**Supplementary Table S18.** csv: addressing method for the relative-protein table.

| property | value |
|---|---|
| name | csv |
| type | Content |
| authors | vega |
| groundable | true |
| description | The two conditions (uninduced, induced_72h), held in the `values` property. |
| addressing_method | row=&col=. |

**Supplementary Table S19.** vega_data_myc_rnaseq_v1: RNA-seq expression.

| property | value |
|---|---|
| name | vega_data_myc_rnaseq_v1 |
| type | Data |
| published_by | @vega |
| authors | vega |
| title | RNA-seq expression, A549 MYC-inducible line, 72h timepoint |
| description | RNA-seq expression (TPM), induced (72h, 1ug/mL doxycycline) and uninduced samples, same passage as the western blot. 19,046 genes quantified, MYC among them. Synthetic example. |

**Supplementary Table S20.** rest: addressing method — query by gene symbol, content not embedded.

| property | value |
| --- | --- |
| name | rest |
| type | Content |
| authors | vega |
| groundable | true |
| description | TPM for the induced and uninduced samples, one value pair per gene. |
| addressing_method | gene=&col= |
| location | community-store://vega_rnaseq_a549_myc_2026 (example locator only, does not resolve to a file) |
| access_method | REST query against the community file store's dataset endpoint, by gene symbol. |

**Supplementary Table S21.** vega_arg_a549_pilot_v1: Argument for taking A549 into the MYC-loss screen pilot.

| property | value |
| --- | --- |
| name | vega_arg_a549_pilot_v1 |
| type | Argument |
| published_by | @vega |
| authors | vega |
| primary_assertion | A0 |
| title | A549 for the MYC-loss modifier screen, taken into the pilot |
| description | MYC is among the most frequently activated drivers in human carcinoma and is not directly druggable, so the laboratory is pursuing its dependencies genetically. The planned screen removes MYC from a cancer line, knocks out each of the remaining ~19,000 genes in a pooled library, and compares guide abundance between MYC-intact and MYC-depleted arms. Guides lost only from the depleted arm mark genes |

| property | value |
|---|---|
| | required to tolerate MYC loss; guides enriched only there mark genes through which the arrest is executed. The screen is informative only in a line that depends on MYC, and resolves hits only if MYC removal slows proliferation without arresting it, since dropout is driven by population doublings and a non-dividing arm cannot be held at library coverage. The line is therefore chosen before any measurement exists in it. One Argument is prepared per candidate line. This one covers A549, a lung adenocarcinoma line. Synthetic example: the measurements and the cited study are invented. |
| purpose | Whether A549 is taken into the pilot. The pilot is a 14-day proliferation assay run under screen conditions, with the library transduced and puromycin selection complete, comparing doubling rate between MYC-intact and MYC-depleted arms. It returns the ratio of the two rates, which sets cell numbers, screen duration and dropout threshold for the full screen. An incorrect choice costs the pilot: two weeks of one person's time and one plate of cells. The pilot is also where the error becomes apparent, and no further commitment is made until it reports. The decision to run the screen is separate and is not addressed here. It commits a genome-wide library, a sequencing run and approximately three months of one person's work, and it is made on the pilot's ratio. |
| verdict | Sufficient for the stated purpose. A549 should be taken into the pilot. The two in-house measurements establish that the inducible cut depletes MYC protein in this line, so the pilot would assay a perturbation that works rather than one that failed. MYC dependence in A549 itself has not been tested. The support is a published result in four other lung adenocarcinoma lines, accepted as reported, together with no known exception for this line. That is an expectation derived from the tumor type rather than a measurement in A549, and it remains to be established directly. The verdict does not extend to running the screen. The screen's design requires the ratio of MYC-depleted to MYC-intact proliferation rate, and that ratio has not been measured in A549 or in any other candidate. The claim should also not be restated as though a proliferation defect had been observed in A549. |
| rationale | Two conditions must hold before the pilot is warranted: A549 must depend on MYC, and the inducible cut must remove MYC in this line. The second is supported locally. At 72h after induction, MYC protein is below the floor of the blot and below the lowest point of the standard curve, and MYC transcript is at 0.11 of parental. These two readouts are not independent. A frameshift at the cut site introduces a premature |

| property | value |
|---|---|
| | termination codon, nonsense-mediated decay degrades the transcript, and protein falls with it, so any successful cut lowers both and their concordance carries little information. The blot is the measurement relied on here. RNA-seq does not report reading frame, and a monoallelic in-frame indel would yield the same transcript value with protein retained. The transcript value is cited as a consistency check and should be read as one. The first is not supported locally. The source study reports proliferative arrest on MYC withdrawal in four lung adenocarcinoma lines and states the result for the tumor type rather than for those four lines, and it is taken here at the scope its authors gave it. Extending it to A549 requires that no exception be known for this line, which is declared as an assumption rather than argued, since it is a negative claim about both the literature and this line and neither has been searched systematically. If that assumption is not granted, nothing in this Argument bears on A549 and the verdict does not stand. The appropriate response would be to pilot two candidate lines in parallel, at roughly twice the cost stated above. The magnitude of the proliferation defect is not addressed in this Argument. It is the parameter the screen design consumes and the quantity the pilot is intended to produce. |

**Supplementary Table S22.** A0: primary claim — A549 depends on MYC, and the cut depletes it.

| property | value |
|---|---|
| name | A0 |
| type | Assertion |
| claim | A549 depends on MYC for proliferation, and the inducible cut depletes MYC protein in this line. |
| scope | Conditions of a MYC-loss modifier screen: A549 in 2D culture, carrying constitutive Cas9 and a doxycycline-inducible MYC sgRNA cassette. Depletion is asserted at 72h in 1ug/mL doxycycline, measured in the parental inducible line before library transduction, and only to the sensitivity of the quantitative blot, whose floor is 0.1 of parental signal: MYC is asserted to be below that floor, not absent. MYC dependence is asserted for proliferation in 2D culture and not for spheroid, soft agar or xenograft growth. No claim is made about the magnitude of the proliferation defect. |

**Supplementary Table S23.** A1: MYC removal slows A549 proliferation (no magnitude asserted).

| property | value |
| --- | --- |
| name | A1 |
| type | Assertion |
| claim | Removing MYC from A549 reduces its proliferation rate. |
| scope | A549 in 2D culture. Qualitative: a reduction is asserted, no magnitude. |

**Supplementary Table S24.** A2: the inducible cut depletes MYC protein in A549.

| property | value |
| --- | --- |
| name | A2 |
| type | Assertion |
| claim | Inducing the MYC sgRNA in A549 reduces MYC protein below the detection floor of the quantitative blot. |
| scope | A549 carrying constitutive Cas9 and the inducible MYC sgRNA cassette, 72h in 1ug/mL doxycycline, parental line before library transduction. Asserted for the bulk population; per-cell editing efficiency is not assessed. |

**Supplementary Table S25.** A3: lung adenocarcinoma lines require MYC for proliferation.

| property | value |
| --- | --- |
| name | A3 |
| type | Assertion |
| claim | Lung adenocarcinoma cell lines require MYC for sustained proliferation, and withdrawing MYC arrests them. |
| scope | Established lung adenocarcinoma lines in 2D culture. Stated as a property of the tumor type, as the source study states it, and not as an enumeration of the lines tested. Individual lines may be exceptions. |

**Supplementary Table S26.** A4: A549 is not a known exception to that generic.

| property | value |
|---|---|
| name | A4 |
| type | Assertion |
| claim | A549 is not an exception to MYC dependence among lung adenocarcinoma lines. |
| scope | A549 as held in this laboratory, within the passage range in use. |

**Supplementary Table S27.** Z1: no known exception for A549, in the literature or in-house.

| property | value |
|---|---|
| name | Z1 |
| type | Assumption |
| rationale | The premise is that no published or in-house result identifies A549 as MYC-independent. No evidence is offered because the claim is a negative. No systematic search of the literature has been made for MYC dependence in A549, and no in-house measurement of proliferation after MYC withdrawal in this line exists. What can be stated is that nothing in this laboratory's records reports A549 as an exception. The premise should be granted as a statement about current coverage rather than about the field. A fitness measurement for MYC in A549 would bear on it directly, and a Member who imports one should expect this Argument to be revised rather than defended. |

**Supplementary Table S28.** G1: western blot — MYC protein below the assay floor.

| property | value |
|---|---|
| name | G1 |
| type | Ground |
| citation | @vega_data_myc_relative_protein_v1.values#csv.row=induced_72h&col=below_floor |

| property | value |
|---|---|
| rationale | Quantitative blot of MYC protein at 72h after induction, densitometry normalized to the tubulin loading control and converted to relative protein by the standard curve. The value falls below the floor of 0.1 of parental signal and below the lowest calibration point of the curve, at 0.25 of the loading range. The measurement establishes that MYC protein was reduced; it does not quantify what remains, and any value between zero and 0.1 of parental is consistent with it. Five percent of parental MYC is not assumed to be functionally inert. |
| criterion | A band at or near parental intensity at 72h would have shown that the cut had not worked, and would have refuted the assertion. The converse is weaker. The antibody is raised against the N-terminal transactivation domain, and MYC has a shorter form initiated at internal codons that lacks that domain while retaining the bHLHZip region for DNA binding and MAX dimerization. A frameshift followed by reinitiation downstream would produce such a product, which is not inert and would not be detected on this blot. Absence of signal establishes absence of the N-terminal epitope, not absence of MYC protein. |

**Supplementary Table S29.** G2: RNA-seq, mechanistically coupled to the blot and weak on its own.

| property | value |
|---|---|
| name | G2 |
| type | Ground |
| citation | @vega_data_myc_rnaseq_v1#rest.gene=MYC&col=TPM_relative |
| rationale | MYC transcript at 0.11 of parental by RNA-seq, from the same passage as the blot. Transcript reduction is the expected consequence of a frameshift: the premature termination codon triggers nonsense-mediated decay, which degrades the message. The value is consistent with disruption of the locus. It is weak support for the assertion. Read counts do not report reading frame. The residual 11% is a mixture of unedited alleles, in-frame indels and edited transcripts whose premature stop escaped decay, and RNA-seq does not separate them. A monoallelic in-frame indel, or reinitiation downstream of the premature stop, would give this same transcript value with functional protein retained. It is also not independent of the blot. Nonsense-mediated decay is the mechanism coupling the two readouts, so any successful cut lowers both, and their concordance adds nothing beyond what the blot already provides. It is cited as a consistency check rather than as evidence of protein loss. |

**Supplementary Table S30.** G3: testimony — lyra's reading of the source study.

| property | value |
|---|---|
| name | G3 |
| type | Ground |
| citation | @lyra_arg_myc_adenocarcinoma_reading_v1.A0 |
| rationale | The primary Assertion of lyra's reading of the source study, taken as it stands. What is taken is that author's assessment of the study rather than the study's abstract: they read the withdrawal timecourses line by line and judged that the arrest is reported for the tumor type and not only for the four lines assayed. That reading is accepted here and is not re-derived. Their verdict comes with the claim. They judged the evidence adequate for nominating a MYC dependency to investigate and inadequate for any statement about MYC inhibition in patients. The present purpose falls inside the first and outside the second, so the verdict is taken as given rather than contested. This is testimony and it is the only support A3 has. The four lines assayed are cited within that Argument and are not counted again here. |

**Supplementary Table S31.** vega_arg_a549_pilot_v1: claim-map relationships.

| rel | source | target |
|---|---|---|
| depends_on | A0 | A1 |
| depends_on | A0 | A2 |
| depends_on | A1 | A3 |
| depends_on | A1 | A4 |
| grounded_by | A2 | G1 |
| grounded_by | A2 | G2 |
| grounded_by | A3 | G3 |
| assumes | A4 | Z1 |